\documentclass{article}

\usepackage[preprint]{corl_2026} % Use for an arXiv preprint; shows authors and removes the CoRL footnote.
\usepackage[utf8]{inputenc} % allow utf-8 input
\usepackage[T1]{fontenc}    % use 8-bit T1 fonts
\usepackage{url}            % simple URL typesetting
\usepackage{nicefrac}       % compact symbols for 1/2, etc.

\hypersetup{
  pdftitle={HAF: Adapting Generalist VLAs to Humanoid Whole-Body Loco-manipulation via Hierarchical Action Flow and Spectral Latent RL},
  pdfauthor={Langzhe Gu, Chengkai Hou, Meng Li, et al.},
  pdfsubject={}
}

\usepackage{microtype}
\usepackage{graphicx}
\usepackage{subcaption}
\usepackage{booktabs} % for professional tables
\usepackage{marvosym}
\usepackage{hyperref}

\usepackage{amsmath}
\usepackage{amssymb}
\usepackage{mathtools}
\usepackage{amsthm}
\usepackage{multirow}
\usepackage{makecell}
\usepackage{xspace}
\usepackage{algorithm}
\usepackage{algpseudocode}
\usepackage{cleveref}

\crefname{figure}{Figure}{Figures}
\Crefname{figure}{Figure}{Figures}

\crefname{table}{Table}{Tables}
\Crefname{table}{Table}{Tables}
\usepackage{wrapfig}

\usepackage{tabularx} % 用于自适应宽度的表格
\usepackage{ragged2e} % 用于单元格内左对齐且换行

\newcommand{\ours}[0]{{HAF}\xspace}
\newcommand{\vla}[0]{{HAF-VLA}\xspace}
\newcommand{\rl}[0]{{HAF-Steer}\xspace}
\title{HAF: Adapting Generalist VLAs to Humanoid Whole-Body Loco-manipulation via Hierarchical Action Flow and Spectral Latent RL}

\author{
\textbf{Langzhe Gu}\textsuperscript{1,2*},
\textbf{Chengkai Hou}\textsuperscript{1,2*},
\textbf{Meng Li}\textsuperscript{2*},
\textbf{Xinhua Wang}\textsuperscript{2},
\textbf{Jiaming Liu}\textsuperscript{1},
\textbf{Xinyuan Lv}\textsuperscript{2,3},
\\
\textbf{Bowei Zhang}\textsuperscript{2,3},
\textbf{Shuanghao Bai}\textsuperscript{2,4},
\textbf{Guangrun Li}\textsuperscript{1,2},
\textbf{Jingyang He}\textsuperscript{1,2},
\textbf{Gaole Dai}\textsuperscript{1},
\textbf{Ziluo Ding}\textsuperscript{2},
\\
\textbf{Zhiyuan Xu}\textsuperscript{2},
\textbf{Kuan Cheng}\textsuperscript{1},
\textbf{Jian Tang}\textsuperscript{2\Letter},
\textbf{Zhengping Che}\textsuperscript{2\Letter},
\textbf{Shanghang Zhang}\textsuperscript{1\Letter}\\[0.5em]
\textsuperscript{1}State Key Laboratory of Multimedia Information Processing, School of Computer Science,\\
Peking University 
\textsuperscript{2}Beijing Innovation Center of Humanoid Robotics 
\textsuperscript{3}Nankai University\\
\textsuperscript{4}Xi’an Jiaotong University
\\[0.3em]
\textsuperscript{*}Equal contribution.
\qquad
\textsuperscript{\Letter}Corresponding author.
}

\begin{document}
\maketitle

%===============================================================================
\vspace{-2em}

\begin{abstract}

Humanoid robots hold great promise as general-purpose agents in human-centered environments, yet generalist vision-language-action (VLA) foundation models are not readily applicable to humanoid whole-body loco-manipulation. The high dimensionality and interdependence of humanoid motions make it challenging for conventional single-stage VLA architectures to coordinate locomotion, waist posture, and dual-arm manipulation effectively. Moreover, policies trained through offline behavior cloning can remain suboptimal during real-world deployment. Although online reinforcement learning can refine policies through real-world interaction, directly tuning large VLA backbones demands excessive computation and may introduce safety risks during real-robot exploration. To address these bottlenecks, we introduce HAF (Humanoid Adaptation Framework), a two-part framework consisting of HAF-VLA and HAF-Steer that transfers off-the-shelf generalist VLA foundation models to humanoid whole-body loco-manipulation. HAF-VLA is a hierarchical action-flow generator built on a pretrained flow-matching VLA. It splits full-body action denoising into three sequential stages with stage embeddings and cross-stage KV caches that retain kinematic dependencies, avoiding incoherent whole-body actions from one-shot generation. On top of the frozen HAF-VLA, HAF-Steer is a latent offline-to-online RL pipeline that leverages flow-matching invertibility and DCT-based dimensionality reduction to restrict RL optimization to a compact noise subspace and train a regularized SAC policy. This avoids updating the large VLA backbone and enables efficient real-world policy refinement. Evaluated on seven real-world humanoid loco-manipulation tasks, HAF surpasses vanilla single-stage VLA baselines and improves whole-body coordination and task performance. Project website: \url{https://grange007.github.io/HAF}.

\end{abstract}


%===============================================================================

%\vspace{-2em}
\section{Introduction}
	
Humanoid robots hold the promise of becoming universal agents capable of operating in diverse human-centric environments. Unlike fixed-base or wheeled manipulators, humanoids require the synchronized coordination of locomotion, torso posture, and bimanual manipulation. However, this complexity introduces a critical challenge: unstable base movements often induce erratic upper-body compensatory motions, which severely degrade balance and manipulation accuracy~\cite{homie,falcon,softa,almi,exbody2,beyondmimic,ze2025generalizable,jiang2026wholebodyvla}.

\begin{figure}[htbp!]
  \vspace{-2em}
  \centering
  \includegraphics[width=0.97\textwidth]{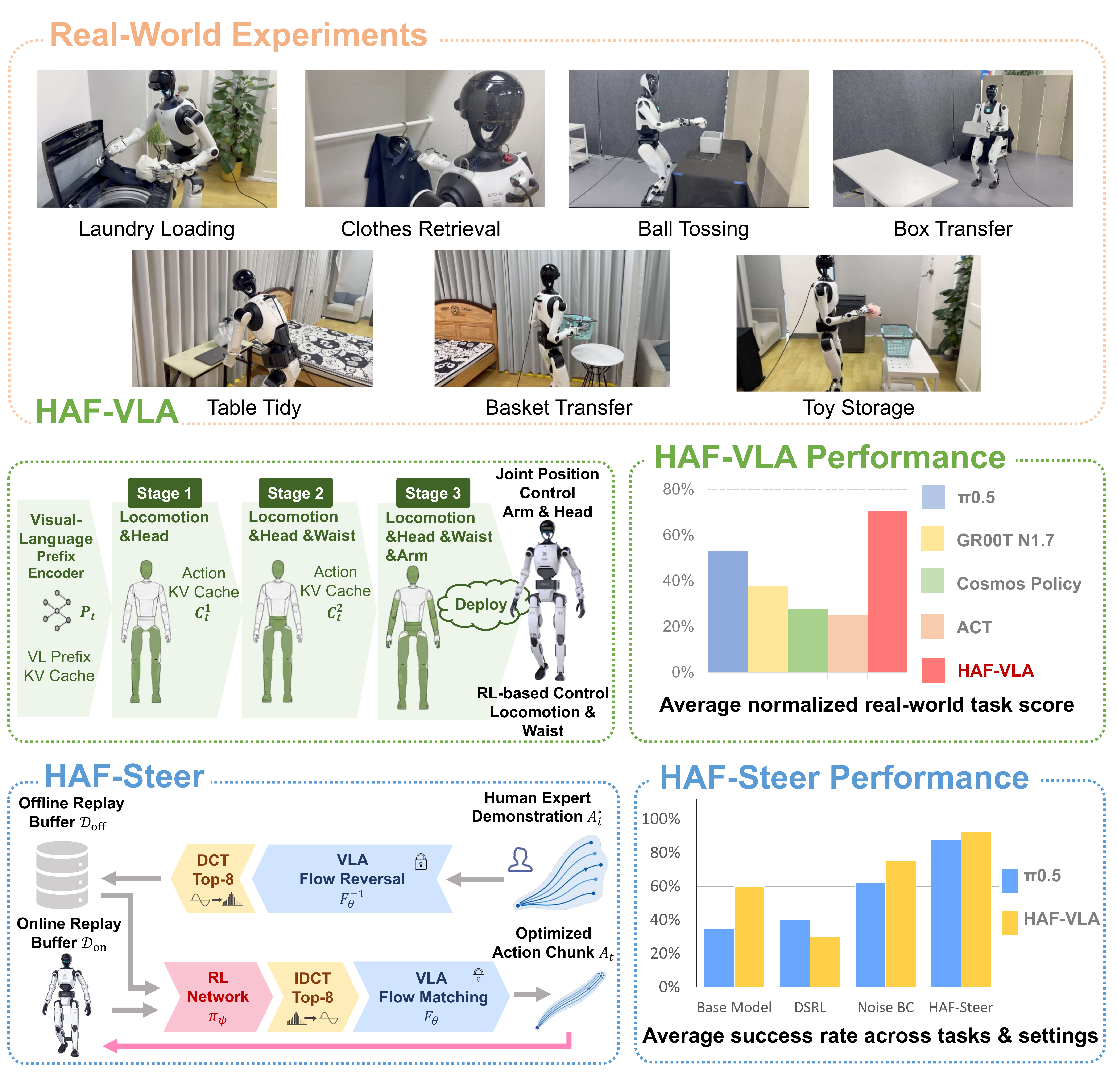}
  \caption{\textbf{Overview of \ours.} We demonstrate seven humanoid loco-manipulation tasks in real-world settings. HAF consists of two complementary components: HAF-VLA is a hierarchical action-flow vision-language-action model that generates whole-body motions stage by stage according to the kinematic dependencies among locomotion, torso adjustment, and dual-arm manipulation; \rl performs latent-space reinforcement learning on discrete cosine transform compressed temporal noise from frozen flow-based VLA policies. Together, they improve task performance in real-world settings.}
  \label{fig:method_overview}
\end{figure}

Although recent generalist vision-language-action (VLA) models have shown strong generalization on conventional robotic platforms~\cite{pi0,pi05,gr00t,kim2026cosmos}, adapting them to humanoid whole-body loco-manipulation remains challenging. Standard flow-matching VLAs typically generate all body-part actions in a single stage, without explicitly modeling the dependencies among locomotion, posture, and manipulation~\cite{beyondmimic,exbody2,omnih2o,r2s2}. Existing humanoid VLAs often address this issue through humanoid-specific pretraining or large embodiment-specific datasets~\cite{jiang2026wholebodyvla,gr00t,wei2026psi0openfoundationmodel}, which substantially increases the cost of adapting a generalist VLA to a new humanoid platform.

Moreover, offline behavior-cloned policies often degrade under deployment-time distribution shifts, yet directly fine-tuning large VLA backbones for high-dimensional humanoid action spaces with online reinforcement learning is computationally expensive and may induce unsafe exploration. Existing latent-noise RL methods avoid backbone updates but either optimize the full high-dimensional temporal noise or repeat a single noise vector over the entire horizon, sacrificing either efficiency or temporal expressiveness~\cite{wagenmaker2025steering,tang2026improving, li2025grrlgoingdexterousprecise, lu2026unisteerunifiednoisesteering}.

To address these challenges, we introduce \textbf{HAF}, the Humanoid Adaptation Framework for transferring pretrained flow-matching VLAs to humanoid whole-body loco-manipulation. \textbf{HAF-VLA} restructures whole-body action generation according to the dependency among locomotion, body posture, and manipulation, while \textbf{HAF-Steer} performs lightweight policy refinement in a compact latent noise space. The two components operate at complementary levels: HAF-VLA provides a structured whole-body action generator, while HAF-Steer further refines the frozen generator through real-world offline-to-online adaptation.

\textbf{HAF-VLA} repurposes a pretrained generalist VLA~\cite{pi05} by orchestrating a progressive, kinematics-aware generation process. Instead of producing full-body actions in a single shot, HAF-VLA sequentially outputs commands for locomotion and head orientation, followed by torso adjustment, and finally bimanual manipulation. This hierarchical ordering prioritizes base stabilization to mitigate the erratic compensatory motions often observed in upper-body control~\cite{chen2025acdit, jiang2025behaviorrobotsuitestreamlining}. To ensure global coherence, we employ cross-stage KV-cache conditioning: clean actions from earlier stages are re-encoded into cross-stage KV caches to condition subsequent generation. Since the active action sets are cumulative, later stages can refine previously generated dimensions, and only the final full-body action chunk is executed.

Built upon the frozen VLA model, \textbf{HAF-Steer} introduces a low-dimensional spectral action space for offline-to-online reinforcement learning (RL). We numerically reverse the frozen flow matching field to recover the initial noise corresponding to demonstrated action chunks and apply the discrete cosine transform (DCT) along the temporal dimension, retaining only the first 8 coefficients. This design avoids the high computational cost of optimizing in the raw high-dimensional action space or fine-tuning the massive VLA backbone, while preserving smoothly varying temporal modes. An RL actor for spectral action is trained through behavior cloning (BC) with offline expert spectral samples first and subsequently optimized using mixed offline--online Soft Actor-Critic (SAC)~\cite{haarnoja2018softactorcriticoffpolicymaximum} with BC regularization.

We validate HAF on two physical humanoid robots across seven complex household loco-manipulation tasks. HAF-VLA significantly outperforms imitation learning and standard VLA baselines, particularly in tasks requiring long-distance travel and coordinated full-body motion. With the HAF-Steer module, our framework achieves robust performance gains in both in-distribution and challenging OOD scenarios.
Our contributions are summarized as follows:
\begin{itemize}
    \item \textbf{Humanoid Adaptation Framework:} We propose \textbf{HAF}, an effective framework that repurposes pretrained generalist VLAs for humanoid whole-body loco-manipulation, eliminating the need to train humanoid-specific foundation models from scratch.
    
    \item \textbf{Kinematics-Aligned Generation:} We design \textbf{HAF-VLA}, a hierarchical action-flow module that aligns with humanoid kinematics to suppress unstable compensatory movements and improve action coherence in whole-body loco-manipulation.

    \item \textbf{Efficient Real-World Refinement:} We develop \textbf{HAF-Steer}, a DCT-based latent RL pipeline that enables stable mixed offline--online adaptation for frozen VLA backbones.
    
    \item \textbf{Comprehensive Real-World Validation:} We validate HAF on two humanoid platforms across seven real-world loco-manipulation tasks, demonstrating improved task performance, whole-body coordination, and robustness under distribution shifts.
\end{itemize}

%===============================================================================

\section{Related Work}
\textbf{Humanoid VLAs.}
Vision-language-action models (VLAs) aim to map visual observations and language instructions directly to robot actions~\cite{rdt, kim2025openvla, zheng2026xvla, univla}. Representative works such as RT-2~\cite{zitkovich2023rt2} and OpenVLA~\cite{kim2025openvla} show that vision-language pretraining can be effectively transferred to robotic manipulation, while $\pi_0$ and $\pi_{0.5}$ further move toward continuous generative action modeling with flow matching~\cite{pi0, pi05}. However, most existing VLAs still focus on tabletop tasks or mobile manipulators, leaving high-dimensional humanoid whole-body action generation underexplored.
Recent works have begun to extend VLAs to humanoid robots~\cite{shao2025langwbclanguagedirectedhumanoidwholebody, xue2025leverb, humanoidvla, bai2026hexhumanoidalignedexpertscrossembodiment}. For instance, GR00T N1.7~\cite{gr00t} studies general-purpose humanoid action generation, WholeBodyVLA~\cite{jiang2026wholebodyvla} explores latent VLA learning for large-space mobile manipulation, and $\Psi_0$~\cite{wei2026psi0openfoundationmodel} introduces an open foundation model pretrained on egocentric human videos. In contrast to these methods that rely on latent skills, cross-embodiment transfer, or learning from human videos, our method is directly built upon a pretrained generalist VLA. We improve its performance on humanoid robots by generating structured whole-body actions grounded in the forward-kinematic chain, sparing the computational cost of training a new model.

\textbf{RL post-training for VLA models.}  Pre-trained VLA policies obtained solely through offline behavior cloning can remain suboptimal during real-world deployment, motivating reinforcement learning (RL) for post-training refinement~\cite{ball2023efficient,zhang2023cherry,lei2025rl,luo2024serl,yang2024robot, li2025grrlgoingdexterousprecise}. Relevant approaches can be broadly grouped into three categories. The first directly fine-tunes the entire VLA backbone using algorithms like PPO~\cite{lei2025rl,mcallister2025flow,ren2025diffusion}, but this incurs massive computational overhead and risks generating unstable, unsafe motions that erase pre-trained priors in high-dimensional whole-body tasks. The second freezes the VLA backbone and trains lightweight residual networks to revise raw actions (e.g., ResFit~\cite{ankile2025residual}, EXPO~\cite{dong2025expo}, DICE-RL~\cite{sun2026prior}). Despite variance reduction techniques, these methods still optimize in the original action space, leading to low sample efficiency and accumulated temporal errors in long-horizon tasks. The third branch performs RL within the latent noise space, which is most relevant to our work. However, existing methods also have some limitations: DSRL~\cite{wagenmaker2025steering} and FRS-based noise policies~\cite{tang2026improving} reduce the search dimension by repeating a single noise vector across the entire action chunk, but this construction departs largely from the independently sampled Gaussian temporal noise used during VLA pretraining. UniSteer~\cite{lu2026unisteerunifiednoisesteering} directly optimizes full-dimensional temporal noise over the entire action chunk, maintaining a large exploration space. Compared with prior latent RL work, HAF-Steer significantly reduces the search dimension of temporal noise, suppresses unstable high-frequency exploratory movements during real-world interaction, and enables more efficient real-world adaptation on long-horizon humanoid loco-manipulation tasks.

\textbf{Whole-body manipulation.}
 Recent research on humanoid whole-body control lays an essential technical foundation for integrated locomotion and manipulation tasks. Advanced low-level controllers and RL-based motion pipelines enable legged robots to perform agile, dynamic movements~\cite{luo2025sonic, li2025bfmzeropromptablebehavioralfoundation, beyondmimic, peng2018deepmimic}. Language-conditioned whole-body controllers such as LangWBC~\cite{shao2025langwbclanguagedirectedhumanoidwholebody} and LeVERB~\cite{xue2025leverb} support high-level language-guided navigation but lack fine-grained bimanual dexterous manipulation capabilities. VR teleoperation frameworks including TWIST2~\cite{ze2025twist2scalableportableholistic}, AMO~\cite{li2025amoadaptivemotionoptimization} and SONIC~\cite{luo2025sonic} deliver effective pipelines to collect full-body humanoid trajectories and optimize low-level motion tracking, yet they merely serve as data acquisition tools rather than generalizable vision-language-driven policies for long-horizon loco-manipulation. None of these whole-body control architectures are integrated with hierarchical generative VLA backbones.

%\label{sec:citations}

	% Citations can be made using either \textbackslash citep\{\} or \textbackslash citet\{\}, depending from the appropriateness. To avoid the citation moving to the next line, it is often a good practice to replace the space before with a tilde (\~{}) character.
	% Example 1: ``CoRL is the best conference ever~\citep{Gauss1857}.''
	% Example 2: ``\citet{Lagrange1788} proved, both theoretically and numerically, that CoRL is the best conference ever.''
	
%===============================================================================

\section{\vla: Hierarchical Action Flow for Humanoid Whole-Body Loco-Manipulation}
\label{sec:haf_vla}

\vla is a hierarchical action-flow module developed to repurpose pretrained generalist flow-matching VLA foundation models for humanoid whole-body loco-manipulation. Unlike single-stage VLA generators that entangle all body movements, it reflects humanoid kinematic dependencies by producing motions step by step: locomotion and head signals are predicted first, followed by waist adjustment, and finally fine bimanual manipulation trajectories. To link motion outputs across generation phases, we introduce a cross-stage KV-cache conditioning mechanism that feeds encoded action features from earlier stages as context to subsequent generation branches, unifying shared visual-linguistic representations and reducing unstable upper-body compensatory movements triggered by flawed base or torso poses.

\begin{figure}[t]
    \centering
    \includegraphics[width=1.0\textwidth]{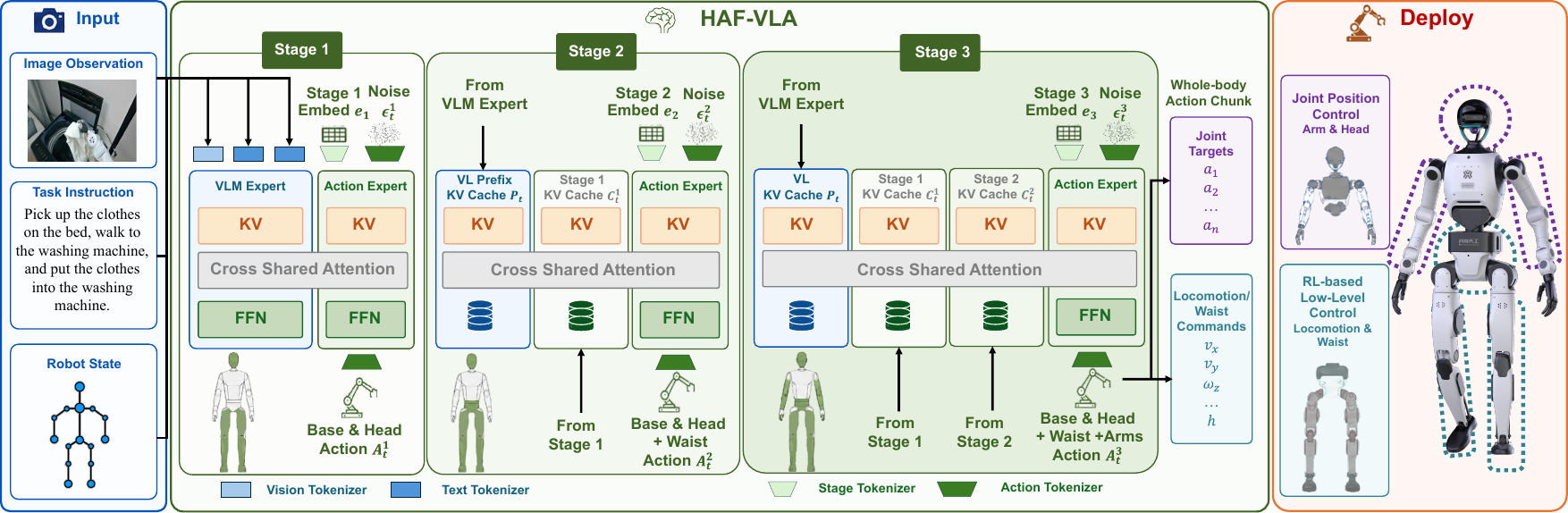}
    \caption{\textbf{Overview of \vla.} A shared action expert progressively
    expands the active action space from locomotion and head control to waist
    posture and bimanual manipulation. Clean action caches from earlier stages
    condition subsequent generation, and only the final full-body action chunk
    is executed.}
    \label{fig:pipeline}
\end{figure}

\subsection{Hierarchical Whole-Body Action Generation}
\label{sec:hierarchical_action_flow}

Conditioned on the current observation $o_t = (I_t, q_t)$, which consists of the egocentric RGB image $I_t$ and robot proprioception $q_t$, as well as the language instruction $\ell$, the VLA model predicts a structured whole-body action chunk $A_t$. Formally, the action chunk is defined as a sequence of future control steps: $A_t = [a_t, \ldots, a_{t+H-1}] \in \mathbb{R}^{H \times D}$, where $H$ denotes the predicted action horizon length and $D$ denotes the dimension of per-step whole-body action. In our implementation, $H=100$ and the robot executes the first 40 steps before the next inference call. To support hierarchical action generation for whole-body loco-manipulation, each instantaneous action $a_t \in \mathbb{R}^D$ is explicitly decomposed into four orthogonal kinematic sub-components: $a_t = [a_t^{\mathrm{move}}, a_t^{\mathrm{head}}, a_t^{\mathrm{waist}}, a_t^{\mathrm{manip}}]$, which correspond to locomotion and skill-mode commands, head orientation control, waist posture adjustment, and bimanual manipulation targets.

Based on this explicit action decomposition, we further partition the full $D$-dimensional action index space into four mutually disjoint subsets corresponding to the above kinematic components: $\mathcal{A}_t^{\mathrm{move}}$, $\mathcal{A}_t^{\mathrm{head}}$, $\mathcal{A}_t^{\mathrm{waist}}$, and $\mathcal{A}_t^{\mathrm{manip}}$. Instead of predicting all action dimensions simultaneously in a single forward pass, HAF-VLA achieves progressive, coarse-to-fine whole-body generation via three nested cumulative action index sets constructed from the four subsets:
\begin{equation}
    \mathcal{A}_t^1 = \mathcal{A}_t^{\mathrm{move}} \cup \mathcal{A}_t^{\mathrm{head}},
    \qquad
    \mathcal{A}_t^2 = \mathcal{A}_t^1 \cup \mathcal{A}_t^{\mathrm{waist}},
    \qquad
      \mathcal{A}_t^3 = \mathcal{A}_t^2 \cup \mathcal{A}_t^{\mathrm{manip}}.
    \label{eq:hierarchical_action_sets}
\end{equation}
This construction strictly establishes the nested inclusion relation $\mathcal{A}_t^1 \subset \mathcal{A}_t^2 \subset \mathcal{A}_t^3$, supporting hierarchical action generation across three progressive stages. In Stage 1, the model generates fundamental locomotion behaviors, task skill modes, and gaze directions over the minimal action set $\mathcal{A}_t^1$. Stage 2 expands the active action space to $\mathcal{A}_t^2$, incorporating waist actuation signals to refine torso posture and reshape the upper-body workspace. Stage 3 fully activates the complete action space $\mathcal{A}_t^3$ to output the final executable whole-body action commands. Different from rigid decoupled generation schemes that fix early-stage outputs, our cumulative nested design allows each subsequent stage to refine and update action dimensions predicted in earlier stages, enabling more accurate and coherent whole-body loco-manipulation.

All three hierarchical stages share identical weights for the action-flow expert network. To reduce redundant computation, the vision-language prefix KV cache is computed only once at the start and reused across all stages of hierarchical inference:
$P_t = F_{\mathrm{prefix}}(o_t,\ell)$, where $P_t$ is the shared vision-language prefix cache derived from the current observation $o_t$ and language instruction $\ell$.
Each stage starts sampling from an independent Gaussian noise vector $\epsilon_t^s\sim\mathcal{N}(\mathbf{0},\mathbf{I})$, and stage-specific behavior is modulated via a trainable stage embedding $e_s$ for stage index $s\in\{1,2,3\}$.
The full multi-stage generation process is formalized below:
\begin{equation}
\begin{aligned}
     A_t^1 &= F_\theta(\epsilon_t^1\mid P_t,e_1),
    & C_t^1 &= \operatorname{Cache}_\theta(A_t^1\mid P_t,e_1),
    \\
     A_t^2 &= F_\theta(\epsilon_t^2\mid P_t,C_t^1,e_2),
    & C_t^2 &= \operatorname{Cache}_\theta(A_t^2\mid P_t,C_t^1,e_2),
    \\
     A_t^3 &= F_\theta(\epsilon_t^3\mid P_t,C_t^1,C_t^2,e_3).
\end{aligned}
\label{eq:explicit_hierarchical_generation}
\end{equation}
Here $F_\theta$ denotes the shared numerical action-flow map, with
the stage-specific action mask applied internally according to the
corresponding stage.
The $\operatorname{Cache}_\theta(\cdot)$ operator re-encodes the denoised action prediction from the current stage and compresses it into an action KV cache ($C_t^1$ for Stage 1, $C_t^2$ for Stage 2). These cached motion embeddings inject coarse prior motion context into later generation steps. Critically, only the final output chunk $A_t \equiv A_t^3$ is sent to the robot controller for real execution; intermediate predictions $A_t^1$ and $A_t^2$ serve purely as context providers and are never deployed directly.

\subsection{Training Strategy and Deployment}
\label{sec:haf_vla_training}

Let $m_s \in \{0,1\}^{D}$ be the binary indicator of the active action set $\mathcal{A}_t^s$, broadcast along the temporal dimension. For compactness, we denote the conditioning of the three stages as
$h_t^1=(P_t,e_1)$,
$h_t^2=(P_t,C_t^1,e_2)$, and
$h_t^3=(P_t,C_t^1,C_t^2,e_3)$.
Given a clean action chunk $A_t\in\mathbb{R}^{H\times D}$, each Stage $s$ independently samples Gaussian noise
$\epsilon_t^s\sim\mathcal{N}(0,I)$ and flow time $\tau^s\sim\mathcal{U}(0,1)$.
The noisy input and target velocity for Stage $s$ are
\begin{equation}
\begin{aligned}
X_{\tau^s,t}^{s}
&=
\left[(1-\tau^s)\epsilon_t^s+\tau^s A_t\right]\odot m_s,\\
u_t^{s}
&=
\left(A_t-\epsilon_t^s\right)\odot m_s,
\end{aligned}
\label{eq:haf_flow_input}
\end{equation}
where $X_{\tau^s,t}^{s}$ is the masked noisy action input and $u_t^{s}$ is the corresponding masked flow-matching target. All stages operate in the same global action space. The stage mask is applied to both the input and target, such that inactive dimensions are assigned zero target velocity rather than excluded from optimization.

The shared action expert is trained with the stage-wise flow-matching objective
\begin{equation}
\mathcal{L}_{\mathrm{HAF}}
=
\mathbb{E}_{A_t,\{\epsilon_t^s,\tau^s\}_{s=1}^{3}}
\left[
\sum_{s=1}^{3}
\frac{1}{HD}
\left\|
v_\theta
\left(
        X_{\tau^s,t}^{s},\tau^s;h_t^{s}
\right)
-
u_t^{s}
\right\|_{F}^{2}
\right],
\label{eq:haf_training}
\end{equation}
where $v_\theta$ denotes the velocity field predicted by the shared action expert. The three stage losses are summed with equal weights. Apart from their independently sampled noise and flow time, the stages differ only in their conditioning $h_t^s$ and active action mask $m_s$, while sharing the same model parameters.

During training, teacher forcing is used to compute cross-stage caches $C_t^1$ and $C_t^2$ from re-encoded masked ground-truth actions rather than stage outputs, which prevents error accumulation while retaining the hierarchical inference structure. At inference, the three stages run sequentially with independent Gaussian noise samples. $C_t^1$ and $C_t^2$ are built from the denoised outputs of earlier stages to condition later ones, and only Stage~3's result is executed via receding-horizon control. Each stage starts from independently sampled noise and performs 10 flow-denoising steps conditioned on the visual-language cache and the available cross-stage action caches.

We deploy HAF-VLA with a receding-horizon execution scheme. The robot executes the first 40
steps before the next inference call. In our real-robot deployment, the complete three-stage inference takes approximately 0.12 seconds on a single RTX 5090 GPU, supporting real-time closed-loop
humanoid loco-manipulation.
% Flow integration steps, execution horizon, and runtime are detailed in experiments.

\section{\rl: Spectral Latent RL for Flow-Matching VLAs}
\label{sec:haf_steer}

\begin{figure*}[htbp]
    \centering
    \includegraphics[width=\textwidth]{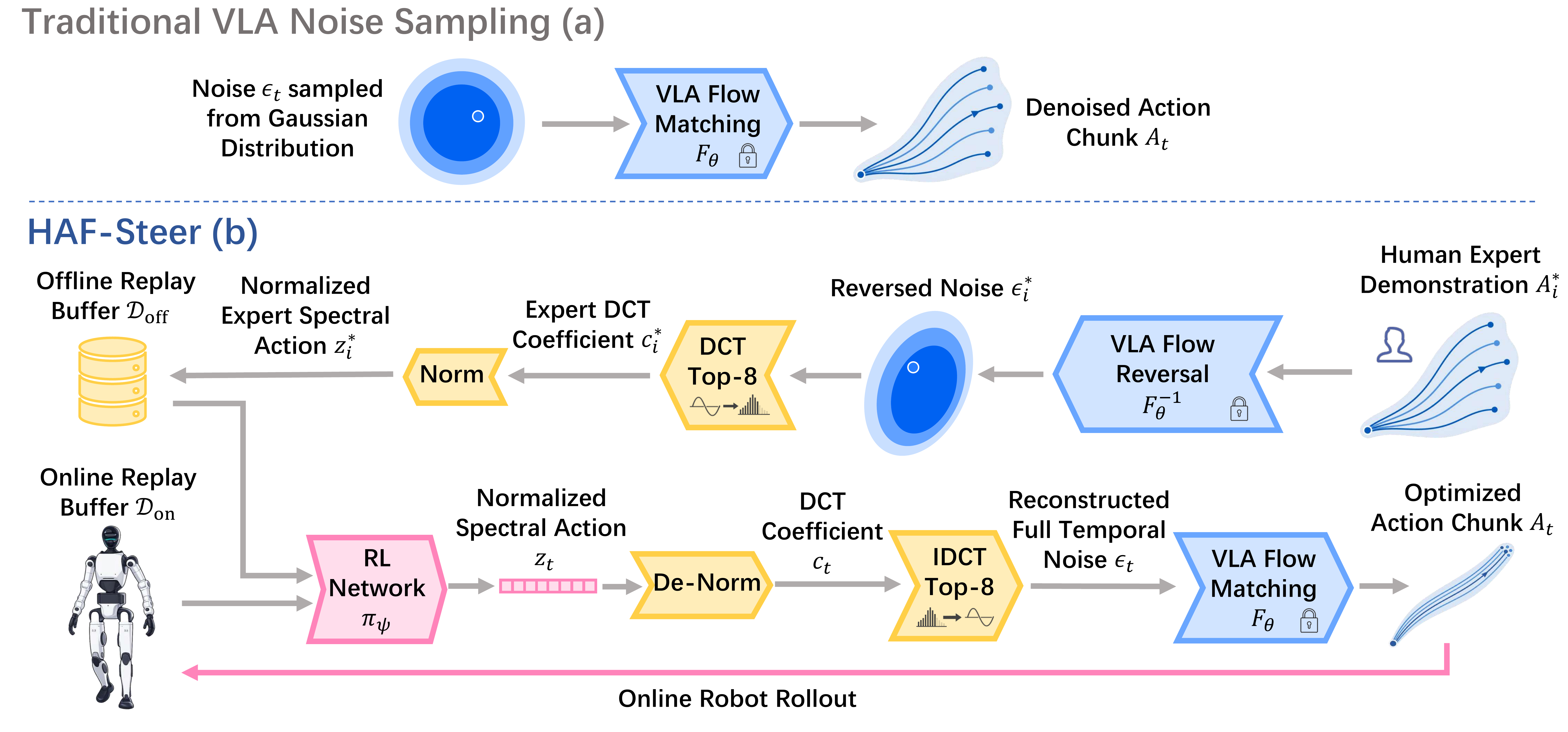}
    \caption{\textbf{Overview of HAF-Steer.}
\textbf{(a)} A conventional flow-matching VLA samples full-dimensional Gaussian noise
and decodes it into an action trajectory through a frozen flow generator.
\textbf{(b)} HAF-Steer instead constructs a low-dimensional spectral action space from
expert demonstrations: demonstrated trajectories are mapped back to their
initial flow noise through numerical reverse integration, compressed by
retaining the first 8 temporal DCT coefficients, and normalized to form
expert spectral actions in the offline replay buffer. An RL policy
is trained with mixed offline and online experience to predict spectral
actions, which are de-normalized, reconstructed into full temporal noise by
inverse DCT, and decoded by the frozen VLA to produce executable robot
trajectories.}
    \label{fig:haf_steer}
\end{figure*}

\rl is a lightweight reinforcement-learning method for adapting a
frozen flow-matching VLA through its initial flow noise, as illustrated in~\cref{fig:haf_steer}. Demonstrated
actions are first mapped back to their corresponding noise and
compressed into low-dimensional spectral actions. A stochastic policy
is then trained in this spectral space and decoded through the frozen
flow generator to produce executable action chunks. We first define
the spectral latent space and then introduce its offline-to-online
optimization.

\subsection{Spectral Latent Space via Flow Reversal}
\label{sec:spectral_latent_space}

HAF-Steer builds upon a frozen pre-trained flow-matching VLA backbone $F_\theta$, which generates whole-body action chunks via a deterministic flow mapping:
$
    A_t = F_\theta\left(\epsilon_t \mid \zeta_t\right), \quad \epsilon_t \in \mathbb{R}^{H\times D},
$    
where $H$ is the action chunk length, $D$ is the dimension of robot action, $A_t\in\mathbb{R}^{H\times D}$ denotes the predicted action chunk, $\epsilon_t$ is the initial flow noise, and $\zeta_t$ represents the unified vision-language-robot conditioning of the frozen VLA model. The network parameters $\theta$ are fully fixed during subsequent reinforcement learning adaptation.

Directly optimizing raw temporal noise at all $H$ time steps leads to an excessively high-dimensional exploration space. Prior steering methods~\cite{wagenmaker2025steering, tang2026improving} repeat a single noise vector over the entire chunk, which we found unstable and frequently inducing severe body jerking or unsafe transient behaviors in whole-body RL training. To enforce temporal smoothness and compress action space complexity, we constrain RL exploration within a low-dimensional spectral subspace spanned by the first $K$ discrete cosine transform (DCT) bases. In our practice, we empirically choose $K=8$, which greatly reduces the exploration space from $H\times D$ to $K\times D$. Our multi-mode spectral parameterization retains diverse smooth temporal variation patterns and yields physically plausible whole-body action corrections.

Based on the above spectral dimensionality reduction design, we further build offline supervised spectral targets to guide policy training. Specifically, to construct supervised spectral targets from offline demonstrations, we first invert each expert action chunk back to its corresponding flow noise space using the frozen VLA model. Given a demonstrated action $A_i^*$ and its conditioning $\zeta_i^*$, we recover and normalize its spectral representation via DCT transformation:
\begin{equation}
    \epsilon_i^* = F_\theta^{-1}\left(A_i^* \mid \zeta_i^*\right),\quad
    c_i^* = \operatorname{DCT}_K(\epsilon_i^*),\quad
    z_i^* = \frac{c_i^*-\mu_c}{\sigma_c+\delta}.
    \label{eq:expert_spectral_action}
\end{equation}
Here, $F_\theta^{-1}$ denotes numerical backward integration of the fixed flow field rather than a learned inverse model. $\epsilon_i^*$ is the recovered full-scale flow noise, $c_i^*\in\mathbb{R}^{K\times D}$ truncates the first $K$ DCT temporal coefficients, and $z_i^*$ is the normalized expert spectral action. The statistics $\mu_c,\sigma_c$ are precomputed over the entire demonstration set, and $\delta$ is a small constant for numerical stability.
According to the derived spectral expert representations, we construct an offline spectral replay buffer for policy training. We formulate VLA-level transition tuples consisting of spectral expert targets, sparse task rewards, and episodic termination flags:
\begin{equation}
    \mathcal D_{\mathrm{off}} = \left\{\left(x_i, z_i^*, r_i, x_{i+1}, d_i\right)\right\}_{i=1}^{N-1},
    \label{eq:offline_spectral_dataset}
    \end{equation}
Here, $x_i$ and $x_{i+1}$ denote consecutive observational states, $z_i^*$ is the precomputed spectral expert target, and $d_i$ indicates episode termination. We adopt a sparse reward scheme for demonstration-guided training: terminal transitions of successful trajectories are assigned $r_i=1$, while all intermediate transitions receive $r_i=0$. This simple yet effective reward formulation drives the policy to learn task-aligned spectral noise corrections from offline data.

\subsection{Offline-to-Online Spectral Policy Learning}
\label{sec:offline_to_online_spectral_rl}

Given the spectral action space and offline replay buffer constructed
above, we first train the actor through behavior cloning and then
improve it using mixed offline--online reinforcement learning. After training, the learned spectral actor generates smooth whole-body corrections at inference time through DCT spectral decoding.

 \textbf{Behavior-cloning initialization.}
 Let $\mu_\psi(x)$ denote the mean output of the stochastic spectral actor $\pi_\psi(z\mid x)$. We first initialize the actor by regressing toward the recovered expert spectral actions in $\mathcal{D}_{\mathrm{off}}$:
\begin{equation}
    \mathcal L_{\mathrm{BC}}
    =
    \mathbb E_{(x_i,z_i^*)\sim \mathcal D_{\mathrm{off}}}
    \left[
        \left\|
            \mu_\psi(x_i)-z_i^*
        \right\|_2^2
    \right].
    \label{eq:spectral_bc_objective}
\end{equation}
Only the spectral actor is optimized during this initialization stage;
the VLA backbone and value functions remain unchanged.

\textbf{Mixed Offline–Online Reinforcement Learning.} After BC initialization, the policy interacts with the real robot following the spectral generation pipeline in Eq.~\eqref{eq:spectral_action_generation}, and newly collected transitions are stored in an online replay buffer $\mathcal{D}_{\mathrm{on}}$. We adopt the same sparse terminal reward rule used for offline data: only successful terminal steps receive $r=1$, while all other transitions yield $r=0$.
In each training iteration, we sample mixed mini-batches $\mathcal{B}=\mathcal{B}_{\mathrm{off}}\cup\mathcal{B}_{\mathrm{on}}$ from both buffers. All samples participate in standard SAC policy updates, while BC regularization is only applied to offline expert data to preserve demonstration quality:
\begin{equation}
\begin{aligned}
    \mathcal L_\pi
    ={}&
    \mathbb E_{x_i\sim\mathcal B,\,z_i\sim\pi_\psi(\cdot\mid x_i)}
    \left[
        \alpha \log\pi_\psi(z_i\mid x_i)
        - \min_j Q_{\omega_j}(x_i,z_i)
    \right]
    \\
    &+
    \lambda_{\mathrm{BC}}
    \mathbb E_{(x_i,z_i^*)\sim \mathcal B_{\mathrm{off}}}
    \left[
        \left\|
            \mu_\psi(x_i)-z_i^*
        \right\|_2^2
    \right].
\end{aligned}
    \label{eq:mixed_spectral_actor_loss}
\end{equation}
Here, $\alpha$ denotes the SAC entropy temperature, $\{Q_{\omega_j}\}$ are twin critic networks, and $\lambda_{\mathrm{BC}}$ balances demonstration regularization strength. The critics are updated via standard entropy-regularized SAC Bellman loss using the full mixed mini-batch.
This hybrid training scheme enables offline demonstrations to stabilize early policy learning and online interactions to adapt to real-world deployment distributions. We gradually decay the offline sampling ratio throughout training to shift optimization from demonstration behavior toward real robot experience. Consistent with our adaptation protocol, the entire flow-matching VLA backbone remains frozen; only the spectral actor, critics, target critics, and entropy temperature are updated during training.

\textbf{Inference Pipeline.} At test time, the trained spectral actor $\pi_\psi$ operates purely in the normalized spectral space. Given the current observation $x_t$, the model samples spectral actions, recovers full temporal noise via inverse DCT, and generates refined whole-body action chunks for robot execution:
\begin{equation}
    z_t \sim \pi_\psi(\cdot \mid x_t),\quad
    c_t = \mu_c + (\sigma_c+\delta)\odot z_t,\quad
    \epsilon_t = \operatorname{IDCT}_K(c_t),\quad
    A_t = F_\theta(\epsilon_t \mid \zeta_t)
    \label{eq:spectral_action_generation}
\end{equation}
The recovered noise $\epsilon_t$ is fed into the frozen VLA flow model conditioned on $\zeta_t$ to produce final executable whole-body actions. The $\operatorname{IDCT}_K$ operation zero-pads truncated spectral coefficients to reconstruct smooth full-horizon temporal noise. By restricting RL optimization to the low-dimensional spectral subspace ($K\ll H$), our framework significantly reduces optimization complexity, suppresses high-frequency jitter, and yields robust, physically plausible whole-body loco-manipulation behaviors.

\subsection{Instantiation on Flow-Matching VLA Backbones}
\label{sec:haf_steer_instantiation}

The formulation above is independent of a particular flow-matching VLA
architecture. We apply the same spectral parameterization, flow-reversal
procedure, and offline-to-online objective to both $\pi_{0.5}$ and \vla.

For $\pi_{0.5}$, $F_\theta$ is its native action-flow map and $\zeta_t$ denotes its
standard visual--language and robot-state conditioning. No change to the
pretrained VLA parameters is required.

For \vla, we instantiate the generic flow map only on its final generation stage. Stages~1 and~2 execute normally and provide their clean-action caches, while
\rl replaces only the initial noise of Stage~3. During demonstration
inversion, $C_t^1$ and $C_t^2$ are constructed through teacher forcing;
during deployment, they are obtained from the predicted clean actions of the
first two stages. In both $\pi_{0.5}$ and \vla, the complete VLA backbone
remains frozen.

\section{Experiments}
\label{sec:result}

\vspace{-0.25em}

\begin{wrapfigure}{r}{0.5\textwidth}
  \centering
  \includegraphics[width=0.5\textwidth]{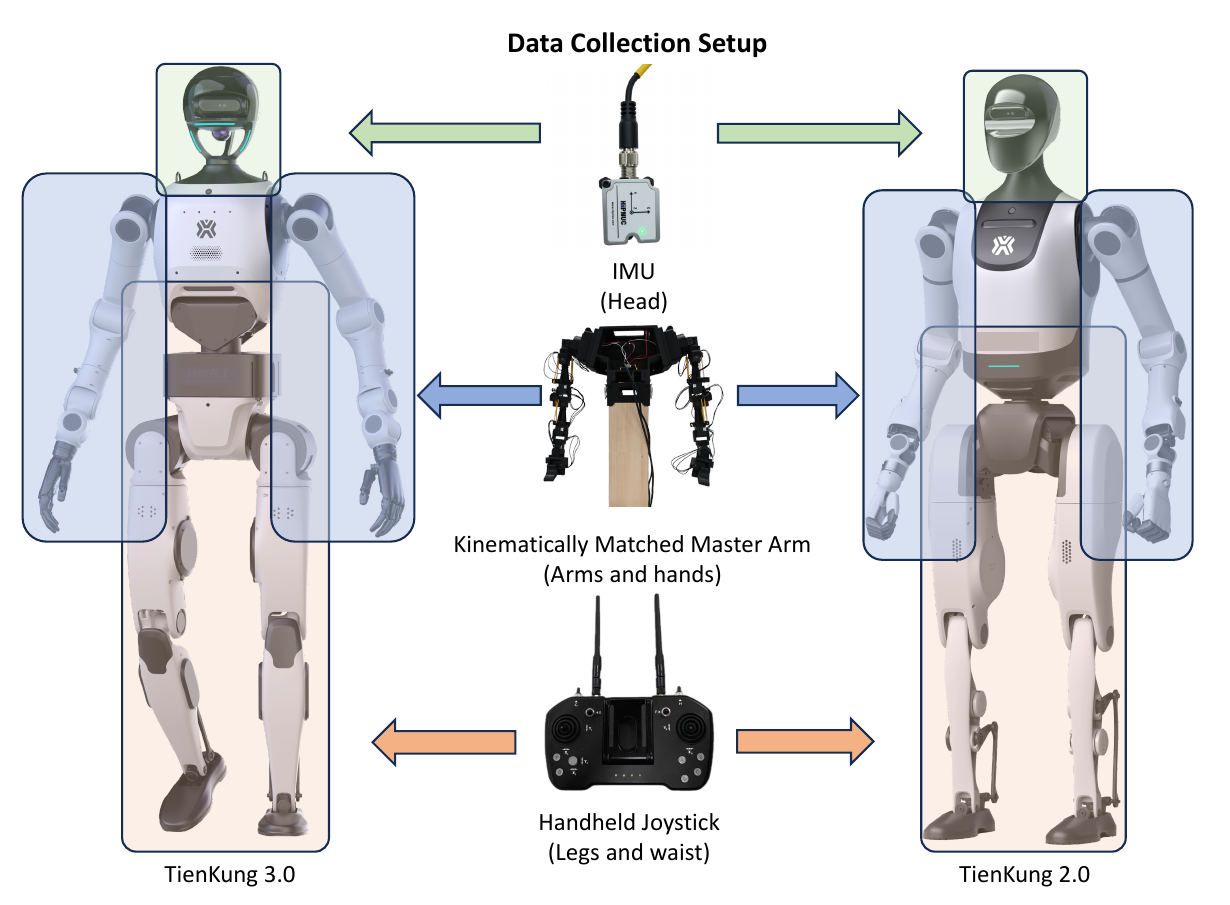}
  \caption{\textbf{Robot teleoperation data collection setup.} The teleoperation pipeline uses an IMU for head control, kinematically matched master arms for dual-arm manipulation, and a joystick to command leg locomotion and waist posture.}
  \label{fig:data_collection}
  \vspace{-2em}
\end{wrapfigure}

In this section, we aim to answer four key research questions: \textbf{Q1.} Does \vla enable long-horizon humanoid loco--manipulation beyond existing state-of-the-art methods? \textbf{Q2.} Does HAF-VLA's hierarchical action-flow design improve imitation-learning performance? \textbf{Q3.} Does \vla generalize to unseen visual and positional disturbances? \textbf{Q4.} Can HAF-Steer improve real-world deployment performance through offline-to-online adaptation?

\subsection{Experiment Setup}

\noindent\textbf{Hardware and Data Collection.} 
We conduct real-robot experiments on the TienKung 2.0 and TienKung 3.0 humanoid platforms. Perception relies on an onboard egocentric RGB camera mounted on the robot head. Training demonstrations are collected via isomorphic teleoperation: a kinematic master arm controls bimanual manipulation, a handheld joystick governs locomotion and waist motion, and an inertial measurement unit (IMU) tracks head orientation commands, as visualized in \cref{fig:data_collection}. We collect 120 teleoperated trajectories for each household task.

\begin{figure}[htbp]
  \centering
  \includegraphics[width=1.0\textwidth]{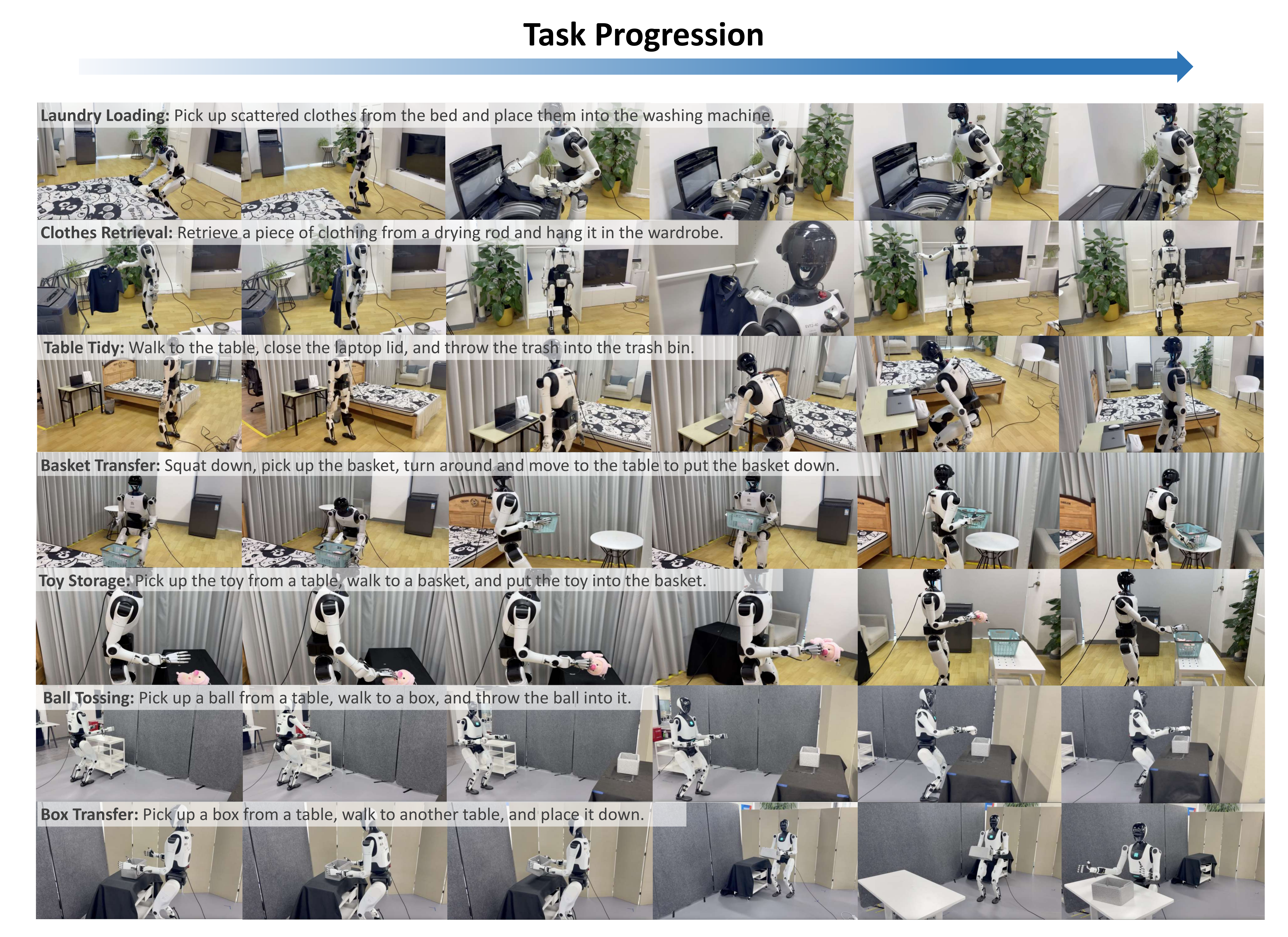}
  \caption{\textbf{Seven real-world humanoid loco-manipulation tasks.} Each row depicts a representative execution sequence requiring navigation, whole-body posture adjustment, and physical object interaction.}
  \label{fig:task_progress}
  \vspace{-1em}
\end{figure}

\noindent\textbf{Tasks and Evaluation Metric.} As illustrated in \cref{fig:task_progress}, we benchmark our framework on seven real-world household loco-manipulation tasks: \textit{Laundry Loading}, \textit{Clothes Retrieval}, \textit{Table Tidy}, \textit{Basket Transfer}, \textit{Toy Storage}, \textit{Ball Tossing}, and \textit{Box Transfer}. These tasks demand long-horizon coordination across locomotion, whole-body posture adjustment, dual-arm manipulation, and object interaction, including challenging skills such as traversing separated regions, torso bending, squatting, object carrying, and throwing. Since binary pass/fail success rates cannot capture partial progress during long sequential executions, we adopt a normalized task score as the primary evaluation metric. Each task is split into predefined milestone sub-goals with individual scores. For each trial rollout, we compute the normalized score as: \[ \mathrm{Score}(\%) = \frac{s_{\mathrm{achieved}}}{s_{\mathrm{max}}} \times 100 . \] We report the average normalized score across 10 independent rollouts per method and task, which enables fine-grained quantitative comparison across tasks with different completion criteria. 

\noindent\textbf{Baselines.} We compare our approach against four representative state-of-the-art baselines: (i) ACT~\cite{act}, a well-established action-chunking imitation learning algorithm; (ii) $\pi_{0.5}$~\cite{pi05}, a strong generalist flow-matching vision-language-action policy; (iii) GR00T N1.7~\cite{gr00t}, a large-scale pretrained humanoid foundation model for general robot skill learning; and (iv) Cosmos Policy~\cite{kim2026cosmos}, a recent world-model-based framework for robotic manipulation. We strictly follow the official open-source implementations and recommended hyperparameters for all baselines.

\subsection{Q1: Does \vla enable long-horizon humanoid loco--manipulation beyond existing SOTA methods?} \label{sec:q1_main_exp} We evaluate \vla across seven long-horizon humanoid loco-manipulation tasks requiring navigation, whole-body coordination, and precise object interaction. As summarized in \cref{tab:vla_score}, \vla achieves the best or tied-best average normalized score on all seven tasks, improving the overall average performance from 53.3\% with the strongest baseline $\pi_{0.5}$ to 70.5\%. Performance gains are particularly pronounced for tasks combining locomotion with subsequent fine manipulation, including \textit{Box Transfer}, \textit{Ball Tossing}, and \textit{Laundry Loading}. Empirically, both $\pi_{0.5}$ and GR00T N1.7 frequently exhibit locomotion drift, while Cosmos Policy is hindered by relatively high online inference latency. Overall, these results demonstrate the effectiveness of hierarchical action generation for high-dimensional humanoid action spaces in whole-body loco-manipulation. \begin{table*}[htbp] \centering \small \setlength{\tabcolsep}{5.5pt} \renewcommand{\arraystretch}{1.15} \caption{ \textbf{Main HAF-VLA results on seven long-horizon humanoid loco--manipulation tasks.} We report the average normalized task score (\%) over 10 rollout trials. Higher values indicate better performance. } \label{tab:vla_score} \begin{tabular}{lccccc} \toprule \textbf{Task} & \textbf{HAF-VLA} & $\boldsymbol{\pi_{0.5}}$ & \textbf{GR00T N1.7} & \textbf{Cosmos Policy} & \textbf{ACT} \\ \midrule Laundry Loading & \textbf{66.7} & 53.3 & 40.0 & 0.0 & 10.0 \\ Clothes Retrieval & \textbf{53.3} & \textbf{53.3} & 33.3 & 26.7 & 23.3 \\ Table Tidy & \textbf{80.0} & 70.0 & 40.0 & 16.7 & 23.3 \\ Basket Transfer & \textbf{63.3} & 50.0 & 43.3 & 33.3 & 16.7 \\ Toy Storage & \textbf{80.0} & 53.3 & 30.0 & 40.0 & 23.3 \\ Ball Tossing & \textbf{56.7} & 33.3 & 36.7 & 3.3 & 30.0 \\ Box Transfer & \textbf{93.3} & 60.0 & 43.3 & 73.3 & 50.0 \\ \midrule \textbf{Average} & \textbf{70.5} & 53.3 & 38.1 & 27.6 & 25.2 \\ \bottomrule \end{tabular} \end{table*}

\subsection{Q2: Does HAF-VLA's hierarchical action-flow design improve imitation-learning performance?} \label{sec:q2_ablation} We conduct targeted ablation experiments on the representative \textit{Laundry Loading} task to isolate the contribution of HAF-VLA's hierarchical action-flow design. Full HAF-VLA progressively expands the active action space across three stages, with 10 flow-matching denoising steps per stage and 30 steps in total. We compare it against three variants: (1) \textit{All-Joint Denoising}, where every stage denoises all action dimensions simultaneously; (2) \textit{Arm-First Hierarchy}, which reverses the coarse-to-fine generation order by prioritizing manipulation before locomotion; and (3) vanilla $\pi_{0.5}$ with 30 denoising steps, which controls for the total denoising budget. \begin{table}[htbp] \vspace{-1em} \centering \small \setlength{\tabcolsep}{4.2pt} \renewcommand{\arraystretch}{1.15} \caption{ \textbf{Ablation study of HAF-VLA's hierarchical action-flow design.} We report the average normalized task score (\%) over 10 rollout trials on the Laundry Loading task. Higher scores are better. } 

\label{tab:ablation_study} 
\begin{tabular}{lccc} 
\toprule \textbf{Method / Variant} & \textbf{Stage Design} & \textbf{Total Steps} & \textbf{Norm. Score (\%)} \\ 
\midrule \textbf{HAF-VLA} & Locomotion/Head $\rightarrow$ + Waist $\rightarrow$ + Manip. & 30 & \textbf{66.7} \\ All-Joint Denoising & All joints at every stage & 30 & 53.3 \\ Arm-First Hierarchy & Manip. $\rightarrow$ + Waist $\rightarrow$ + Locomotion/Head & 30 & 50.0 \\ $\pi_{0.5}$, 30 steps & Full-body action & 30 & 20.0 \\ 
\bottomrule \end{tabular} \end{table} 

As shown in \cref{tab:ablation_study}, our full hierarchical design outperforms all ablated variants, indicating that the improvement does not simply result from increasing the number of denoising iterations. The inferior performance of the arm-first variant further highlights the importance of establishing locomotion and body posture before fine manipulation. We also observe that increasing $\pi_{0.5}$ from 10 to 30 denoising steps decreases performance despite the larger computation budget, which is mainly caused by the increased inference latency (from 0.075s to 0.115s). Since $\pi_{0.5}$ already exhibits locomotion-prediction errors on humanoid tasks, the additional latency can amplify the temporal mismatch between predicted velocity commands and robot execution. In contrast, HAF-VLA remains effective at a similar latency, suggesting greater robustness to inference delay.

\subsection{Q3: Does \vla generalize to unseen visual and positional disturbances?}
\label{sec:q3_exp_generalization}

\begin{figure}[htbp!]
  \centering
  \includegraphics[width=1.0\textwidth]{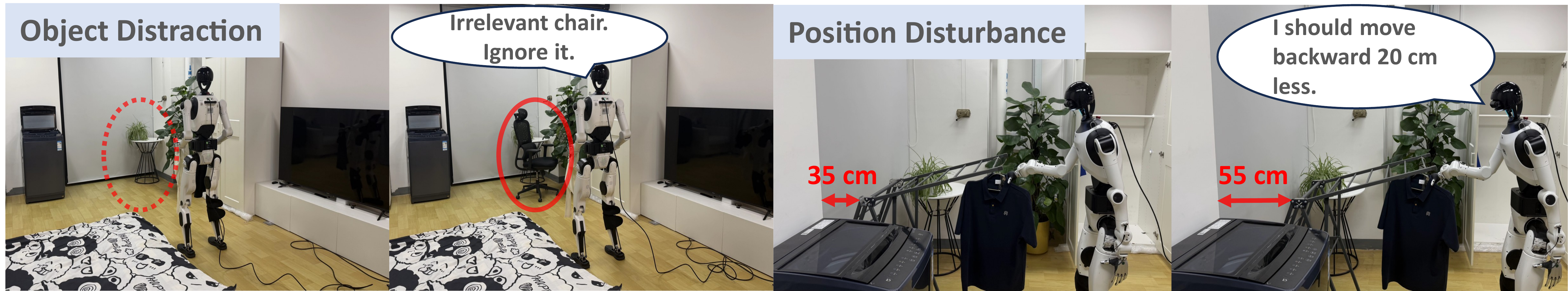}
  \caption{
  \textbf{Generalization experiments.}
  Left: object-distraction test during \textit{Laundry Loading}. 
  Right: position-disturbance test during \textit{Clothes Retrieval}.
  }
  \label{fig:generalization}
\end{figure}

We evaluate out-of-distribution robustness under two controlled distribution shifts visualized in \cref{fig:generalization}.
For \textit{Laundry Loading}, we place an unseen black office chair along the robot navigation path to introduce visual distraction.
For \textit{Clothes Retrieval}, we shift the robot initial position 20 cm backward relative to the demonstration collection setup.
We compare \vla against vanilla $\pi_{0.5}$ using the same normalized task score metric across 10 rollouts per condition.
As reported in \cref{tab:generalization}, HAF-VLA maintains higher task performance under both visual distraction and positional perturbation, which demonstrates stronger generalization beyond the exact demonstration layout.

\begin{table}[htbp]
 \vspace{-1em}
    \centering
    \small
    \setlength{\tabcolsep}{5pt}
    \renewcommand{\arraystretch}{1.15}
    \caption{
    \textbf{Generalization evaluation under controlled disturbances.}
    We report the average normalized task score (\%) over 10 rollout trials. Higher scores indicate stronger robustness.
    }
    \label{tab:generalization}
    \begin{tabular}{llcc}
        \toprule
        \textbf{Task} 
        & \textbf{Disturbance} 
        & \textbf{HAF-VLA} 
        & $\boldsymbol{\pi_{0.5}}$ \\
        \midrule
        Laundry Loading 
        & Unseen chair near the path 
        & \textbf{40.0} 
        & 26.7 \\
        Clothes Retrieval 
        & 20 cm backward start shift 
        & \textbf{43.3} 
        & 36.7 \\
        \bottomrule
    \end{tabular}
    \vspace{-1em}
\end{table}

\label{sec:q4_cross_vla}

\subsection{Q4: Can HAF-Steer improve real-world deployment performance through offline-to-online adaptation?}
\label{sec:rl}

\begin{wrapfigure}{r}{0.6\textwidth}
  \centering
  \vspace{-2em}
  \includegraphics[width=0.6\textwidth]{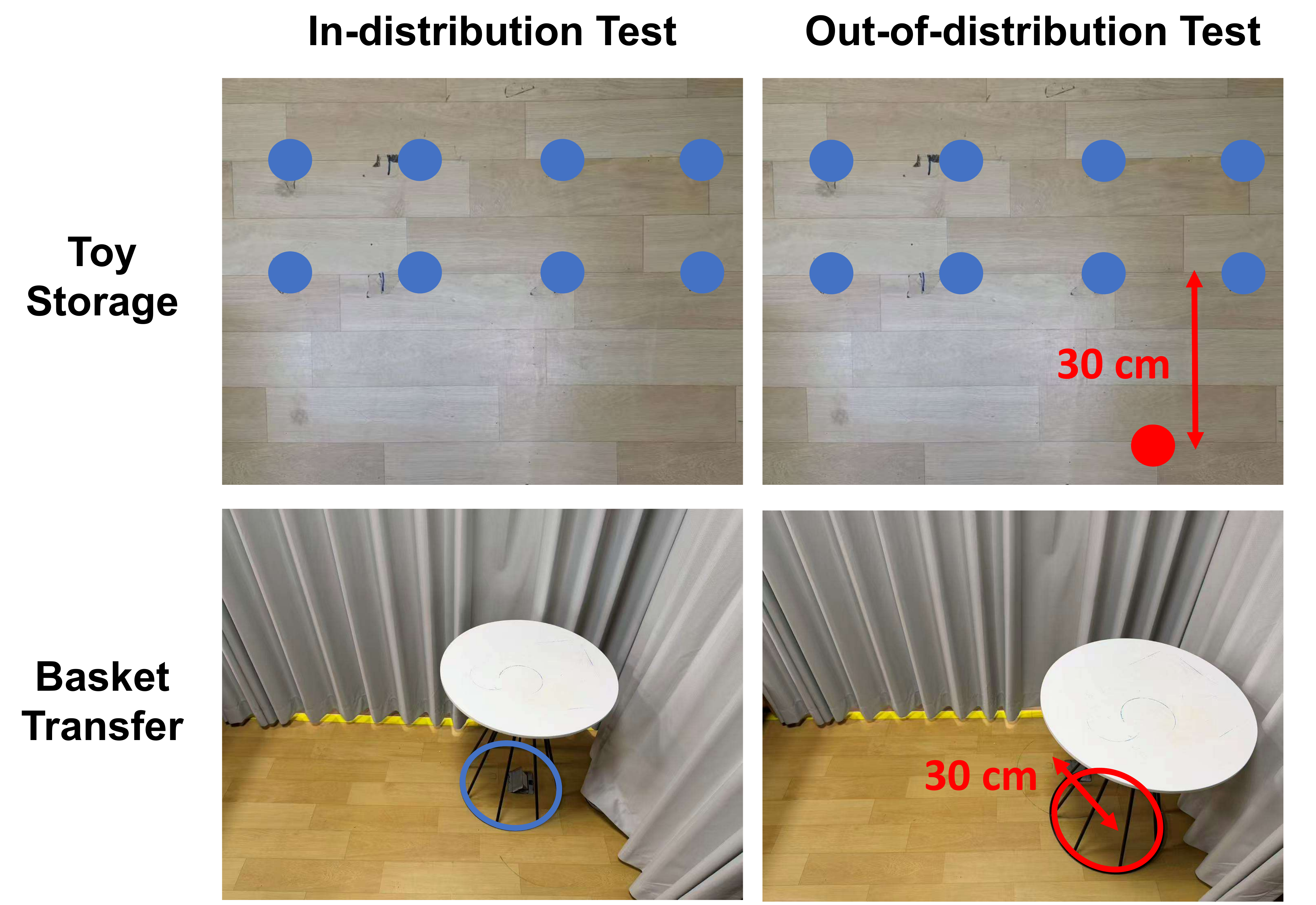}
  \caption{
  \textbf{In-distribution and out-of-distribution evaluation setups for HAF-Steer.}
  Blue markers indicate destination-table locations covered during demonstration collection, while red markers indicate the OOD evaluation locations.
  For \textit{Toy Storage}, the destination table is placed 30\,cm beyond the demonstrated spatial range.
  For \textit{Basket Transfer}, the destination table is shifted by 30\,cm from its demonstrated position.
  }
  \label{fig:rl_ood}
\end{wrapfigure}

We evaluate HAF-Steer on two representative whole-body loco-manipulation tasks, \textit{Toy Storage} and \textit{Basket Transfer}, using the TienKung 3.0 humanoid platform.
For each task, we consider both in-distribution (ID) settings covered by the offline demonstrations and out-of-distribution (OOD) settings with unseen goal locations.
Specifically, for \textit{Toy Storage}, the destination table is moved 30\,cm beyond the spatial range covered during demonstration collection; for \textit{Basket Transfer}, the destination table is similarly shifted by 30\,cm from its demonstrated position.
These controlled spatial perturbations require the policy to adapt its locomotion and whole-body manipulation behavior to previously unseen goal configurations.
To examine whether the proposed spectral adaptation mechanism generalizes across different flow-matching policies, we conduct all experiments on both the vanilla $\pi_{0.5}$ backbone and HAF-VLA.
The corresponding ID and OOD evaluation setups are visualized in \cref{fig:rl_ood}.

\noindent\textbf{Baselines and Variants.}
We compare four policy variants.
\textit{Base Model} directly deploys the frozen VLA without latent adaptation.
\textit{DSRL}~\cite{wagenmaker2025steering} optimizes a single $D$-dimensional noise vector and repeats it across the entire action horizon, thereby reducing the search dimension at the cost of removing temporal variation in the flow noise.
\textit{Noise BC} evaluates our spectral actor immediately after the spectral behavior-cloning initialization, without subsequent SAC training.
Finally, \textit{HAF-Steer} further optimizes the BC-initialized spectral actor using mixed offline--online SAC while keeping the entire VLA backbone frozen.
This comparison isolates both the effect of the spectral noise representation and the additional benefit of online reinforcement learning.

\begin{figure}[htbp!]
  \centering
  \includegraphics[width=1.0\textwidth]{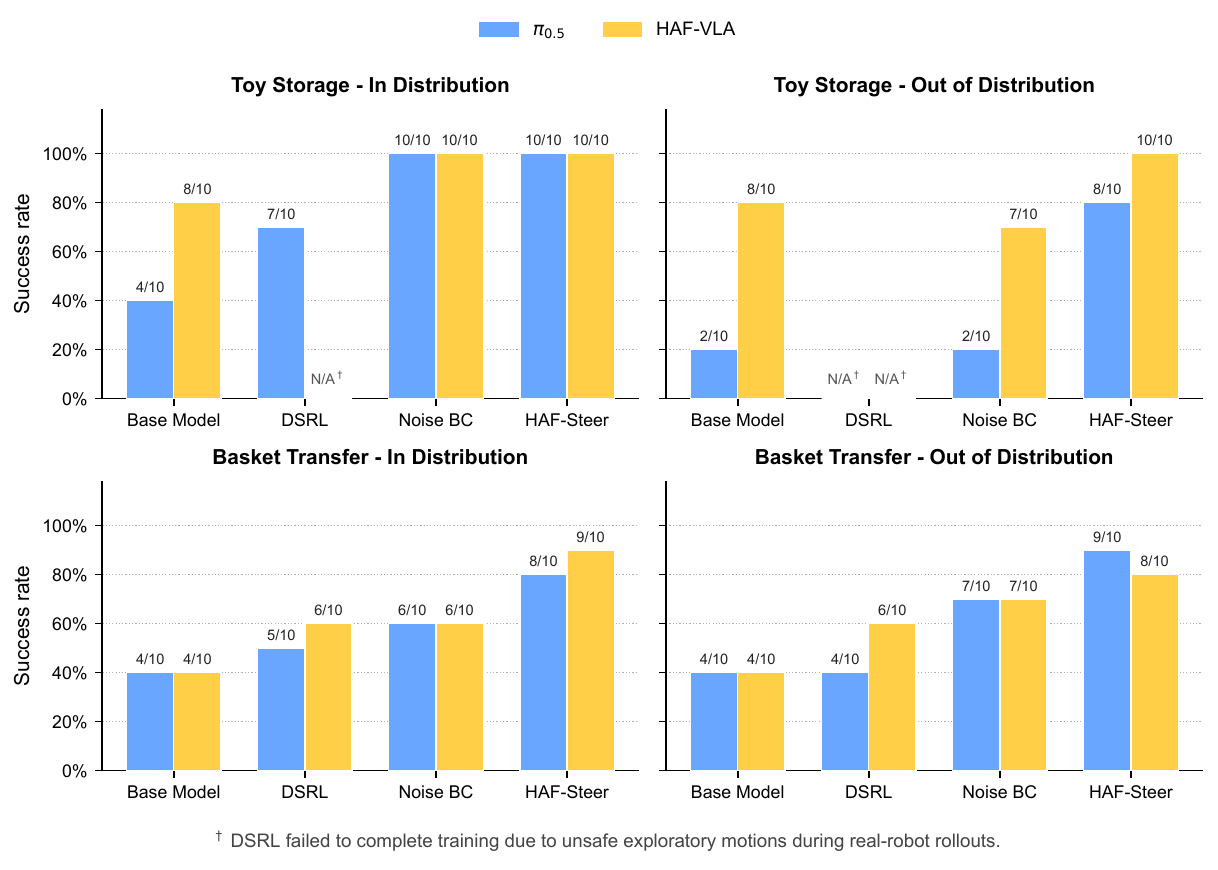}
  \caption{
  \textbf{HAF-Steer performance under in-distribution and out-of-distribution settings.}
  We report successful trials out of 10 rollouts on \textit{Toy Storage} and \textit{Basket Transfer} using both $\pi_{0.5}$ and HAF-VLA backbones.
  Noise BC denotes the spectral actor after behavior-cloning initialization without SAC post-training.
  $^\dagger$ DSRL failed to complete training because unsafe exploratory motions triggered early termination during real-robot rollouts.
  }
  \label{fig:rl_exp}
\end{figure}

\noindent\textbf{Results Analysis.} As shown in \cref{fig:rl_exp}, HAF-Steer consistently improves both $\pi_{0.5}$ and HAF-VLA across ID and OOD settings. Noise BC already improves performance in several settings, while subsequent mixed offline–online RL further increases real-world success rates, demonstrating that HAF-Steer can refine deployment performance beyond the offline-initialized policy in both ID and OOD conditions. In contrast, the repeated-noise parameterization of DSRL removes temporal variation and produces unsafe exploratory motions in several real-robot experiments, leading to early training termination. By retaining multiple low-frequency temporal modes, HAF-Steer enables more stable latent-space exploration and effective online adaptation. 

Importantly, HAF-VLA and HAF-Steer operate at complementary levels. HAF-VLA first provides a structured whole-body policy through hierarchical action generation, while HAF-Steer further adapts this frozen policy to deployment-time distribution shifts. Applying HAF-Steer to HAF-VLA improves its success rate in all four evaluated ID/OOD settings and achieves the best or tied-best performance in three of them. These results demonstrate the benefit of combining structured action generation and lightweight latent policy adaptation within the unified HAF framework.

%===============================================================================

\section{Conclusion and Limitations}
\label{sec:conclusion}

In this paper, we introduce HAF, a unified framework for adapting pretrained VLA models to humanoid whole-body loco-manipulation. It comprises two key components: HAF-VLA generates structured sequential actions using progressive denoising and cross-stage KV-cache conditioning, while HAF-Steer leverages SAC-based latent reinforcement learning to optimize DCT-compressed noise coefficients and improve real-world deployment performance through offline-to-online adaptation. HAF successfully unifies structured generative modeling and adaptive policy optimization for complex humanoid whole-body loco-manipulation.

\textbf{Limitations.} The hierarchical pipeline increases denoising computation and introduces deployment latency. The latent RL module is also constrained by the base VLA’s inherent priors and may therefore fail to correct erroneous motions in extreme unseen scenarios. Future work will explore efficient denoising schemes and lightweight RL implementations to enhance real-time performance and task robustness.

%===============================================================================

\section*{Acknowledgements}
\label{sec:acknowledge}

This work was supported by the National Natural Science Foundation of China (62476011), the Beijing Natural Science Foundation (L252060), and the Beijing Major Science and Technology Project under Contract no. Z191100010618003.

% \clearpage
% The acknowledgments are automatically included only in the final and preprint versions of the paper.
% Add acknowledgments here for the preprint or camera-ready version if needed.

%===============================================================================

\bibliography{main}  % .bib

@misc{li2025bfmzeropromptablebehavioralfoundation,
      title={BFM-Zero: A Promptable Behavioral Foundation Model for Humanoid Control Using Unsupervised Reinforcement Learning}, 
      author={Yitang Li and Zhengyi Luo and Tonghe Zhang and Cunxi Dai and Anssi Kanervisto and Andrea Tirinzoni and Haoyang Weng and Kris Kitani and Mateusz Guzek and Ahmed Touati and Alessandro Lazaric and Matteo Pirotta and Guanya Shi},
      year={2025},
      eprint={2511.04131},
      archivePrefix={arXiv},
      primaryClass={cs.RO},
      url={https://arxiv.org/abs/2511.04131}, 
}

@misc{luo2025sonic,
      title={SONIC: Supersizing Motion Tracking for Natural Humanoid Whole-Body Control}, 
      author={Zhengyi Luo and Ye Yuan and Tingwu Wang and Chenran Li and Fernando Castañeda and Sirui Chen and Zi-Ang Cao and Jiefeng Li and David Minor and Qingwei Ben and Jinhyung Park and David Sami and Zi Wang and Xingye Da and Runyu Ding and Cyrus Hogg and Lina Song and Edy Lim and Eugene Jeong and Tairan He and Haoru Xue and Wenli Xiao and Simon Yuen and Jan Kautz and Yan Chang and Umar Iqbal and Linxi "Jim" Fan and Yuke Zhu},
      year={2026},
      eprint={2511.07820},
      archivePrefix={arXiv},
      primaryClass={cs.RO},
      url={https://arxiv.org/abs/2511.07820}, 
}

@InProceedings{zitkovich2023rt2,
  title = 	 {RT-2: Vision-Language-Action Models Transfer Web Knowledge to Robotic Control},
  author =       {Zitkovich, Brianna and Yu, Tianhe and Xu, Sichun and Xu, Peng and Xiao, Ted and Xia, Fei and Wu, Jialin and Wohlhart, Paul and Welker, Stefan and Wahid, Ayzaan and Vuong, Quan and Vanhoucke, Vincent and Tran, Huong and Soricut, Radu and Singh, Anikait and Singh, Jaspiar and Sermanet, Pierre and Sanketi, Pannag R. and Salazar, Grecia and Ryoo, Michael S. and Reymann, Krista and Rao, Kanishka and Pertsch, Karl and Mordatch, Igor and Michalewski, Henryk and Lu, Yao and Levine, Sergey and Lee, Lisa and Lee, Tsang-Wei Edward and Leal, Isabel and Kuang, Yuheng and Kalashnikov, Dmitry and Julian, Ryan and Joshi, Nikhil J. and Irpan, Alex and Ichter, Brian and Hsu, Jasmine and Herzog, Alexander and Hausman, Karol and Gopalakrishnan, Keerthana and Fu, Chuyuan and Florence, Pete and Finn, Chelsea and Dubey, Kumar Avinava and Driess, Danny and Ding, Tianli and Choromanski, Krzysztof Marcin and Chen, Xi and Chebotar, Yevgen and Carbajal, Justice and Brown, Noah and Brohan, Anthony and Arenas, Montserrat Gonzalez and Han, Kehang},
  booktitle = 	 {Proceedings of The 7th Conference on Robot Learning},
  pages = 	 {2165--2183},
  year = 	 {2023},
  editor = 	 {Tan, Jie and Toussaint, Marc and Darvish, Kourosh},
  volume = 	 {229},
  series = 	 {Proceedings of Machine Learning Research},
  month = 	 {06--09 Nov},
  publisher =    {PMLR},
  url = 	 {https://proceedings.mlr.press/v229/zitkovich23a.html}
}

@InProceedings{kim2025openvla,
  title = 	 {{OpenVLA}: An Open-Source Vision-Language-Action Model},
  author =       {Kim, Moo Jin and Pertsch, Karl and Karamcheti, Siddharth and Xiao, Ted and Balakrishna, Ashwin and Nair, Suraj and Rafailov, Rafael and Foster, Ethan P and Sanketi, Pannag R and Vuong, Quan and Kollar, Thomas and Burchfiel, Benjamin and Tedrake, Russ and Sadigh, Dorsa and Levine, Sergey and Liang, Percy and Finn, Chelsea},
  booktitle = 	 {Proceedings of The 8th Conference on Robot Learning},
  pages = 	 {2679--2713},
  year = 	 {2025},
  editor = 	 {Agrawal, Pulkit and Kroemer, Oliver and Burgard, Wolfram},
  volume = 	 {270},
  series = 	 {Proceedings of Machine Learning Research},
  month = 	 {06--09 Nov},
  publisher =    {PMLR},
  url = 	 {https://proceedings.mlr.press/v270/kim25c.html}
}

@inproceedings{
zheng2026xvla,
title={X-{VLA}: Soft-Prompted Transformer as Scalable Cross-Embodiment Vision-Language-Action Model},
author={Jinliang Zheng and Jianxiong Li and Zhihao Wang and Dongxiu Liu and Xirui Kang and Yuchun Feng and Yinan Zheng and Jiayin Zou and Yilun Chen and Jia Zeng and Tai Wang and Ya-Qin Zhang and Jingjing Liu and Xianyuan Zhan},
booktitle={The Fourteenth International Conference on Learning Representations},
year={2026},
url={https://openreview.net/forum?id=kt51kZH4aG}
}

@article{
	peng2018deepmimic,
	author = {Peng, Xue Bin and Abbeel, Pieter and Levine, Sergey and van de Panne, Michiel},
	title = {DeepMimic: Example-guided Deep Reinforcement Learning of Physics-based Character Skills},
	journal = {ACM Trans. Graph.},
	issue_date = {August 2018},
	volume = {37},
	number = {4},
	month = jul,
	year = {2018},
	issn = {0730-0301},
	pages = {143:1--143:14},
	articleno = {143},
	numpages = {14},
	url = {http://doi.acm.org/10.1145/3197517.3201311},
	doi = {10.1145/3197517.3201311},
	acmid = {3201311},
	publisher = {ACM},
	address = {New York, NY, USA},
}

@article{xue2025leverb,
  author       = {Haoru Xue and
                  Xiaoyu Huang and
                  Dantong Niu and
                  Qiayuan Liao and
                  Thomas Kragerud and
                  Jan Tommy Gravdahl and
                  Xue Bin Peng and
                  Guanya Shi and
                  Trevor Darrell and
                  Koushil Sreenath and
                  S. Shankar Sastry},
  title        = {LeVERB: Humanoid Whole-Body Control with Latent Vision-Language Instruction},
  journal      = {CoRR},
  volume       = {abs/2506.13751},
  year         = {2025},
  url          = {https://doi.org/10.48550/arXiv.2506.13751},
  doi          = {10.48550/ARXIV.2506.13751},
  eprinttype   = {arXiv},
  eprint       = {2506.13751},
  bibsource    = {dblp computer science bibliography, https://dblp.org}
}

@inproceedings{
jiang2026wholebodyvla,
title={WholeBody{VLA}: Towards Unified Latent {VLA} for Whole-body Loco-manipulation Control},
author={Haoran Jiang and Jin Chen and Qingwen Bu and Li Chen and Modi Shi and Yanjie Zhang and Delong Li and Chuanzhe Suo and Chuang Wang and Zhihui Peng and Hongyang Li},
booktitle={The Fourteenth International Conference on Learning Representations},
year={2026},
url={https://openreview.net/forum?id=OCJmVjyzN7}
}

@misc{shao2025langwbclanguagedirectedhumanoidwholebody,
      title={LangWBC: Language-directed Humanoid Whole-Body Control via End-to-end Learning}, 
      author={Yiyang Shao and Xiaoyu Huang and Bike Zhang and Qiayuan Liao and Yuman Gao and Yufeng Chi and Zhongyu Li and Sophia Shao and Koushil Sreenath},
      year={2025},
      eprint={2504.21738},
      archivePrefix={arXiv},
      primaryClass={cs.RO},
      url={https://arxiv.org/abs/2504.21738}, 
}

@article{exbody2,
  title={{ExBody2}: Advanced Expressive Humanoid Whole-Body Control}, 
  author={Ji, Mazeyu and Peng, Xuanbin and Liu, Fangchen and Li, Jialong and Yang, Ge and Cheng, Xuxin and Wang, Xiaolong},
  journal={arXiv preprint arXiv:2412.13196},
  year={2024},
  }

@inproceedings{omnih2o,
  title={{OmniH2O}: Universal and Dexterous Human-to-Humanoid Whole-Body Teleoperation and Learning},
  author={He, Tairan and Luo, Zhengyi and He, Xialin and Xiao, Wenli and Zhang, Chong and Zhang, Weinan and Kitani, Kris M and Liu, Changliu and Shi, Guanya},
  booktitle={Conference on Robot Learning},
  year={2025}
}

@article{beyondmimic,
  title={{BeyondMimic}: From Motion Tracking to Versatile Humanoid Control via Guided Diffusion},
  author={Liao, Qiayuan and Truong, Takara E. and Huang, Xiaoyu and Gao, Yuman and Tevet, Guy and Sreenath, Koushil and Liu, C. Karen},
  journal={arXiv preprint arXiv:2508.08241},
  year={2025}
}

@misc{li2025amoadaptivemotionoptimization,
      title={AMO: Adaptive Motion Optimization for Hyper-Dexterous Humanoid Whole-Body Control}, 
      author={Jialong Li and Xuxin Cheng and Tianshu Huang and Shiqi Yang and Ri-Zhao Qiu and Xiaolong Wang},
      year={2025},
      eprint={2505.03738},
      archivePrefix={arXiv},
      primaryClass={cs.RO},
      url={https://arxiv.org/abs/2505.03738}, 
}

@misc{ze2025twist2scalableportableholistic,
      title={TWIST2: Scalable, Portable, and Holistic Humanoid Data Collection System}, 
      author={Yanjie Ze and Siheng Zhao and Weizhuo Wang and Angjoo Kanazawa and Rocky Duan and Pieter Abbeel and Guanya Shi and Jiajun Wu and C. Karen Liu},
      year={2025},
      eprint={2511.02832},
      archivePrefix={arXiv},
      primaryClass={cs.RO},
      url={https://arxiv.org/abs/2511.02832}, 
}

@misc{homie,
      title={HOMIE: Humanoid Loco-Manipulation with Isomorphic Exoskeleton Cockpit}, 
      author={Qingwei Ben and Feiyu Jia and Jia Zeng and Junting Dong and Dahua Lin and Jiangmiao Pang},
      year={2025},
      eprint={2502.13013},
      archivePrefix={arXiv},
      primaryClass={cs.RO},
      url={https://arxiv.org/abs/2502.13013}, 
}

@article{falcon,
  title={{FALCON}: Learning Force-Adaptive Humanoid Loco-Manipulation},
  author={Zhang, Yuanhang and Yuan, Yifu and Gurunath, Prajwal and He, Tairan and Omidshafiei, Shayegan and Agha-mohammadi, Ali-akbar and Vazquez-Chanlatte, Marcell and Pedersen, Liam and Shi, Guanya},
  journal={arXiv preprint arXiv:2505.06776},
  year={2025}
}

@article{softa,
  title={Learning Gentle Humanoid Locomotion and End-Effector Stabilization Control},
  author={Li, Yitang and Zhang, Yuanhang and Xiao, Wenli and Pan, Chaoyi and Weng, Haoyang and He, Guanqi and He, Tairan and Shi, Guanya},
  journal={arXiv preprint arXiv:2505.24198},
  year={2025}
}

@article{almi,
  title={Adversarial Locomotion and Motion Imitation for Humanoid Policy Learning},
  author={Shi, Jiyuan and Liu, Xinzhe and Wang, Dewei and Lu, Ouyang and Schwertfeger, S{\"o}ren and Sun, Fuchun and Bai, Chenjia and Li, Xuelong},
  journal={arXiv preprint arXiv:2504.14305},
  year={2025}
}

@article{r2s2,
  title={Unleashing Humanoid Reaching Potential via Real-world-Ready Skill Space},
  author={Zhang, Zhikai and Chen, Chao and Xue, Han and Wang, Jilong and Liang, Sikai and Liu, Yun and Zhang, Zongzhang and Wang, He and Yi, Li},
  journal={arXiv preprint arXiv:2505.10918},
  year={2025}
}

@inproceedings{head,
  title={Hand-Eye Autonomous Delivery: Learning Humanoid Navigation, Locomotion and Reaching},
  author={Chen, Sirui and Ye, Yufei and Cao, Zi-Ang and Lew, Jennifer and Xu, Pei and Liu, C Karen},
  booktitle=CORL,
  year={2025}
}

@article{humanoidvla,
  title={{Humanoid-VLA}: Towards universal humanoid control with visual integration},
  author={Ding, Pengxiang and Ma, Jianfei and Tong, Xinyang and Zou, Binghong and Luo, Xinxin and Fan, Yiguo and Wang, Ting and Lu, Hongchao and Mo, Panzhong and Liu, Jinxin and others},
  journal={arXiv preprint arXiv:2502.14795},
  year={2025}
}

@article{gr00t,
  title={{GR00T N1}: An open foundation model for generalist humanoid robots},
  author={{NVIDIA} and Bjorck, Johan and Casta{\~n}eda, Fernando and Cherniadev, Nikita and Da, Xingye and Ding, Runyu and Fan, Linxi and Fang, Yu and Fox, Dieter and Hu, Fengyuan and Huang, Spencer and others},
  journal={arXiv preprint arXiv:2503.14734},
  year={2025}
}

@misc{act,
      title={Learning Fine-Grained Bimanual Manipulation with Low-Cost Hardware}, 
      author={Tony Z. Zhao and Vikash Kumar and Sergey Levine and Chelsea Finn},
      year={2023},
      eprint={2304.13705},
      archivePrefix={arXiv},
      primaryClass={cs.RO},
      url={https://arxiv.org/abs/2304.13705}, 
}

@misc{rdt,
      title={RDT-1B: a Diffusion Foundation Model for Bimanual Manipulation}, 
      author={Songming Liu and Lingxuan Wu and Bangguo Li and Hengkai Tan and Huayu Chen and Zhengyi Wang and Ke Xu and Hang Su and Jun Zhu},
      year={2025},
      eprint={2410.07864},
      archivePrefix={arXiv},
      primaryClass={cs.RO},
      url={https://arxiv.org/abs/2410.07864}, 
}

@article{pi0,
  author       = {Kevin Black and
                  Noah Brown and
                  Danny Driess and
                  Adnan Esmail and
                  Michael Equi and
                  Chelsea Finn and
                  Niccolo Fusai and
                  Lachy Groom and
                  Karol Hausman and
                  Brian Ichter and others},
  title        = {{\(\pi\)}\({}_{\mbox{0}}\): {A} Vision-Language-Action Flow Model
                  for General Robot Control},
  journal      = {arXiv preprint arXiv:2410.24164},
  year         = {2024},
}

@article{pi05,
  title={{\(\pi\)}\({}_{0.5}\): a Vision-Language-Action Model with Open-World Generalization},
  author={{Physical Intelligence} and Black, Kevin and Brown, Noah and Darpinian, James and Dhabalia, Karan and Driess, Danny and Esmail, Adnan and Equi, Michael and Finn, Chelsea and Fusai, Niccolo and others},
  journal={arXiv preprint arXiv:2504.16054},
  year={2025}
}

@misc{univla,
      title={Unified Vision-Language-Action Model}, 
      author={Yuqi Wang and Xinghang Li and Wenxuan Wang and Junbo Zhang and Yingyan Li and Yuntao Chen and Xinlong Wang and Zhaoxiang Zhang},
      year={2025},
      eprint={2506.19850},
      archivePrefix={arXiv},
      primaryClass={cs.CV},
      url={https://arxiv.org/abs/2506.19850}, 
}

@misc{wei2026psi0openfoundationmodel,
      title={$\Psi_0$: An Open Foundation Model Towards Universal Humanoid Loco-Manipulation}, 
      author={Songlin Wei and Hongyi Jing and Boqian Li and Zhenyu Zhao and Jiageng Mao and Zhenhao Ni and Sicheng He and Jie Liu and Xiawei Liu and Kaidi Kang and Sheng Zang and Weiduo Yuan and Marco Pavone and Di Huang and Yue Wang},
      year={2026},
      eprint={2603.12263},
      archivePrefix={arXiv},
      primaryClass={cs.RO},
      url={https://arxiv.org/abs/2603.12263}, 
}

@misc{jiang2025behaviorrobotsuitestreamlining,
      title={BEHAVIOR Robot Suite: Streamlining Real-World Whole-Body Manipulation for Everyday Household Activities}, 
      author={Yunfan Jiang and Ruohan Zhang and Josiah Wong and Chen Wang and Yanjie Ze and Hang Yin and Cem Gokmen and Shuran Song and Jiajun Wu and Li Fei-Fei},
      year={2025},
      eprint={2503.05652},
      archivePrefix={arXiv},
      primaryClass={cs.RO},
      url={https://arxiv.org/abs/2503.05652}, 
}

@inproceedings{
kim2026cosmos,
title={Cosmos Policy: Fine-Tuning Video Models for Visuomotor Control and Planning},
author={Moo Jin Kim and Yihuai Gao and Tsung-Yi Lin and Yen-Chen Lin and Yunhao Ge and Grace Lam and Percy Liang and Shuran Song and Ming-Yu Liu and Chelsea Finn and Jinwei Gu},
booktitle={The Fourteenth International Conference on Learning Representations},
year={2026},
url={https://openreview.net/forum?id=wPEIStHxYH}
}

@misc{li2025grrlgoingdexterousprecise,
      title={GR-RL: Going Dexterous and Precise for Long-Horizon Robotic Manipulation}, 
      author={Yunfei Li and Xiao Ma and Jiafeng Xu and Yu Cui and Zhongren Cui and Zhigang Han and Liqun Huang and Tao Kong and Yuxiao Liu and Hao Niu and Wanli Peng and Jingchao Qiao and Zeyu Ren and Haixin Shi and Zhi Su and Jiawen Tian and Yuyang Xiao and Shenyu Zhang and Liwei Zheng and Hang Li and Yonghui Wu},
      year={2025},
      eprint={2512.01801},
      archivePrefix={arXiv},
      primaryClass={cs.RO},
      url={https://arxiv.org/abs/2512.01801}, 
}

@misc{lu2026unisteerunifiednoisesteering,
      title={{UniSteer}: Unified Noise Steering for Efficient Human-Guided VLA Adaptation}, 
      author={Junjie Lu and Xinyao Qin and Yuhua Jiang and Kaixin Wang and Chuheng Zhang and Bin Liang and Jun Yang and Min Xu and Li Zhao},
      year={2026},
      eprint={2605.10821},
      archivePrefix={arXiv},
      primaryClass={cs.RO},
      url={https://arxiv.org/abs/2605.10821}, 
}

@misc{chen2025acdit,
      title={AC-DiT: Adaptive Coordination Diffusion Transformer for Mobile Manipulation}, 
      author={Sixiang Chen and Jiaming Liu and Siyuan Qian and Han Jiang and Lily Li and Renrui Zhang and Zhuoyang Liu and Chenyang Gu and Chengkai Hou and Pengwei Wang and Zhongyuan Wang and Shanghang Zhang},
      year={2025},
      eprint={2507.01961},
      archivePrefix={arXiv},
      primaryClass={cs.RO},
      url={https://arxiv.org/abs/2507.01961}, 
}

@inproceedings{ze2025generalizable,
  title={Generalizable humanoid manipulation with 3d diffusion policies},
  author={Ze, Yanjie and Chen, Zixuan and Wang, Wenhao and Chen, Tianyi and He, Xialin and Yuan, Ying and Peng, Xue Bin and Wu, Jiajun},
  booktitle={IROS},
  year={2025},
}

@misc{haarnoja2018softactorcriticoffpolicymaximum,
      title={Soft Actor-Critic: Off-Policy Maximum Entropy Deep Reinforcement Learning with a Stochastic Actor}, 
      author={Tuomas Haarnoja and Aurick Zhou and Pieter Abbeel and Sergey Levine},
      year={2018},
      eprint={1801.01290},
      archivePrefix={arXiv},
      primaryClass={cs.LG},
      url={https://arxiv.org/abs/1801.01290}, 
}

@inproceedings{ball2023efficient,
  title={Efficient online reinforcement learning with offline data},
  author={Ball, Philip J and Smith, Laura and Kostrikov, Ilya and Levine, Sergey},
  booktitle={International Conference on Machine Learning},
  pages={1577--1594},
  year={2023},
  organization={PMLR}
}

@article{zhang2023cherry,
  title={Cherry-picking with reinforcement learning: Robust dynamic grasping in unstable conditions},
  author={Zhang, Yunchu and Ke, Liyiming and Deshpande, Abhay and Gupta, Abhishek and Srinivasa, Siddhartha},
  journal={arXiv preprint arXiv:2303.05508},
  year={2023}
}

@article{lei2025rl,
  title={Rl-100: Performant robotic manipulation with real-world reinforcement learning},
  author={Lei, Kun and Li, Huanyu and Yu, Dongjie and Wei, Zhenyu and Guo, Lingxiao and Jiang, Zhennan and Wang, Ziyu and Liang, Shiyu and Xu, Huazhe},
  journal={arXiv preprint arXiv:2510.14830},
  year={2025}
}

@inproceedings{luo2024serl,
  title={Serl: A software suite for sample-efficient robotic reinforcement learning},
  author={Luo, Jianlan and Hu, Zheyuan and Xu, Charles and Tan, You Liang and Berg, Jacob and Sharma, Archit and Schaal, Stefan and Finn, Chelsea and Gupta, Abhishek and Levine, Sergey},
  booktitle={2024 IEEE International Conference on Robotics and Automation (ICRA)},
  pages={16961--16969},
  year={2024},
  organization={IEEE}
}

@inproceedings{yang2024robot,
  title={Robot fine-tuning made easy: Pre-training rewards and policies for autonomous real-world reinforcement learning},
  author={Yang, Jingyun and Mark, Max Sobol and Vu, Brandon and Sharma, Archit and Bohg, Jeannette and Finn, Chelsea},
  booktitle={2024 IEEE International Conference on Robotics and Automation (ICRA)},
  pages={4804--4811},
  year={2024},
  organization={IEEE}
}

@inproceedings{ren2025diffusion,
  title={Diffusion policy policy optimization},
  author={Ren, Allen and Lidard, Justin and Ankile, Lars and Simeonov, Anthony and Agrawal, Pulkit and Majumdar, Anirudha and Burchfiel, Benjamin and Dai, Hongkai and Simchowitz, Max},
  booktitle={International Conference on Learning Representations},
  volume={2025},
  pages={77288--77329},
  year={2025}
}

@article{mcallister2025flow,
  title={Flow matching policy gradients},
  author={McAllister, David and Ge, Songwei and Yi, Brent and Kim, Chung Min and Weber, Ethan and Choi, Hongsuk and Feng, Haiwen and Kanazawa, Angjoo},
  journal={arXiv preprint arXiv:2507.21053},
  year={2025}
}

@article{ankile2025residual,
  title={Residual off-policy rl for finetuning behavior cloning policies},
  author={Ankile, Lars and Jiang, Zhenyu and Duan, Rocky and Shi, Guanya and Abbeel, Pieter and Nagabandi, Anusha},
  journal={arXiv preprint arXiv:2509.19301},
  year={2025}
}

@article{dong2025expo,
  title={Expo: Stable reinforcement learning with expressive policies},
  author={Dong, Perry and Li, Qiyang and Sadigh, Dorsa and Finn, Chelsea},
  journal={arXiv preprint arXiv:2507.07986},
  year={2025}
}

@article{sun2026prior,
  title={From prior to pro: Efficient skill mastery via distribution contractive rl finetuning},
  author={Sun, Zhanyi and Song, Shuran},
  journal={arXiv preprint arXiv:2603.10263},
  year={2026}
}

@misc{bai2026hexhumanoidalignedexpertscrossembodiment,
      title={HEX: Humanoid-Aligned Experts for Cross-Embodiment Whole-Body Manipulation}, 
      author={Shuanghao Bai and Meng Li and Xinyuan Lv and Jiawei Wang and Xinhua Wang and Fei Liao and Chengkai Hou and Langzhe Gu and Wanqi Zhou and Kun Wu and Ziluo Ding and Zhiyuan Xu and Lei Sun and Shanghang Zhang and Zhengping Che and Jian Tang and Badong Chen},
      year={2026},
      eprint={2604.07993},
      archivePrefix={arXiv},
      primaryClass={cs.RO},
      url={https://arxiv.org/abs/2604.07993}, 
}

@article{wagenmaker2025steering,
  title={Steering your diffusion policy with latent space reinforcement learning},
  author={Wagenmaker, Andrew and Nakamoto, Mitsuhiko and Zhang, Yunchu and Park, Seohong and Yagoub, Waleed and Nagabandi, Anusha and Gupta, Abhishek and Levine, Sergey},
  journal={arXiv preprint arXiv:2506.15799},
  year={2025}
}

@article{tang2026improving,
  title={Improving Robotic Generalist Policies via Flow Reversal Steering},
  author={Tang, Andy and Chen, William and Wagenmaker, Andrew and Finn, Chelsea and Levine, Sergey},
  journal={arXiv preprint arXiv:2606.13675},
  year={2026}
}

\end{document}